\documentclass[10pt, a4paper]{article}

\usepackage[final]{lrec2026} 
\usepackage{booktabs}
\usepackage{amsmath}
\usepackage{tipa}

\usepackage{tcolorbox}
\tcbuselibrary{listings}
\newtcolorbox{promptbox}[1][]{
  colback=gray!5,    
  colframe=gray!50,  
  fonttitle=\bfseries,
  coltitle=black,
  title=#1,
  boxrule=0.5pt,
  arc=2mm,           
  outer arc=2mm,
  left=4pt,right=4pt,top=4pt,bottom=4pt,
  after skip=12pt,
  listing only,
  listing options={
    basicstyle=\ttfamily\small,
    breaklines=true,
    columns=fullflexible
  }
}
\usepackage{array}    
\usepackage{ragged2e} 
\usepackage{amssymb}

\title{\textit{Meenz bleibt Meenz},\\but Large Language Models Do Not Speak Its Dialect}

\name{
Minh Duc Bui$^{\nabla}$ \quad
Manuel Mager$^{\nabla}$\\
{\bf \large Peter Herbert Kann$^{\spadesuit}$ \quad Katharina von der Wense$^{\nabla,\clubsuit}$}
}

\address{
$^{\nabla}$\,Johannes Gutenberg University Mainz, Germany \quad
$^{\spadesuit}$\,Marburg University, Germany \\
$^{\clubsuit}$\,University of Colorado Boulder, USA \\
{\tt minhducbui@uni-mainz.de}
}

\abstract{
\textit{Meenzerisch}, the dialect spoken in the German city of Mainz, is also the traditional language of the Mainz carnival, a yearly celebration well known throughout Germany. However, Meenzerisch is on the verge of dying out---a fate it shares with many other German dialects. Natural language processing (NLP) has the potential to help with the preservation and revival efforts of languages and dialects. However, so far no NLP research has looked at Meenzerisch. 
This work presents the first research in the field of NLP that is explicitly focused on the dialect of Mainz. We introduce a digital dictionary---an NLP-ready dataset derived from an existing resource \cite{schramm1966mainzer}---to support researchers in modeling and benchmarking the language. It contains 2,351 words in the dialect paired with their meanings described in Standard German. We then use this dataset to answer the following research questions: (1) Can state-of-the-art large language models (LLMs) generate definitions for dialect words? (2) Can LLMs generate words in Meenzerisch, given their definitions? Our experiments show that LLMs can do \textit{neither}: the best model for definitions reaches only $6.27\%$ accuracy and the best word generation model's accuracy is $1.51\%$. We then conduct two additional experiments in order to see if accuracy is improved by few-shot learning and by extracting rules from the training set, which are then passed to the LLM. While those approaches are able to improve the results, accuracy remains below $10\%$. This highlights that additional resources and an intensification of research efforts focused on German dialects are desperately needed.
 \\ \newline  }

\begin{document}

\maketitleabstract

\section{Introduction}
Most German dialects are heavily endangered, as speakers face dialect-related discrimination \cite{wirtz2025functional} and the homogenizing effects of top-down language standardization \cite{Rutten_Vosters_2021}. This has contributed to a wider underappreciation of dialects, not only among non-speakers and international audiences, but occasionally among speakers themselves. Yet, language variants are a huge part of each region's cultural identity, and speakers who cherish their local cultural traditions fear the looming loss of their dialect. 

Language technology has the potential to aid the preservation of minority languages, including dialects \cite{galla2016indigenous}. Furthermore, natural language processing (NLP) systems able to handle dialects would allow speakers to interact with contemporary language technology in a natural way \cite{mager-etal-2018-challenges}. However, since dialects are mostly used in speech, little dialectal text data is available. This is why large language models (LLMs) have been shown to perform poorly for many German dialects \cite{kantharuban-etal-2023-quantifying, peng-etal-2024-sebastian, Blaschke_2025, bar-etal-2025-swiss, munoz-ortiz-etal-2025-evaluating}. 

Here, we present---to the best of our knowledge---the first study investigating the ability of LLMs to handle a dialect that has so far been overlooked by the NLP community: the dialect of the city of Mainz---called \textit{Meenzerisch} or \textit{M{\"a}{\"a}nzerisch} by the speakers themselves.\footnote{To explain the title: the English translation of the Meenzer \textit{Meenz bleibt Meenz} is \textit{Mainz remains Mainz}.} This dialect, while being largely unknown outside Germany, is the language of the carnival of Mainz (\textit{Meenzer Fassenacht} in the dialect), and carnival speeches in the dialect of Mainz are broadcast on German national television once a year. Traditionally, those speeches were given by native speakers. However, in recent years, more and more presenters have instead been native speakers of Standard German, highlighting the dialect's risk of decline.

As a first step towards supporting Meenzerisch with NLP, we introduce a novel dataset resulting from semi-automatically digitizing a physical dictionary of Meenzerisch. It contains dialectal words together with a description of their meaning in Standard German. We further use this dataset to investigate the level of understanding of LLMs by tasking them to generate the meaning of the given words. To explore their generation capabilities, we also prompt them for the opposite direction: given a description of a word's meaning, generating the word itself. 

We experiment with a diverse set of open-source LLMs, including small, medium-sized, and large models. Our results show that all evaluated LLMs struggle to generate accurate definitions for the dialect words. On average, models achieve only 4.24\% accuracy, with the best-performing model, Llama-3.3 70B, reaching just 6.27\%. Performance is even lower on the second task---generating the correct dialect word from its meaning---, where the average accuracy drops to 0.56\%, and the strongest model, GPT-OSS 120B, attains only 1.51\%. While few-shot learning and automatic rule extraction slightly improve results for definition generation (up to 9.24\% and 8.37\%, respectively), overall performance remains low.

\paragraph{Contributions} To summarize, our contributions are as follows: 1) the first dataset containing words in the dialect of Mainz together with their definitions; 2) the first study of LLMs' abilities to comprehend words in the dialect of Mainz, showing that current models struggle to understand them; and 3) the first study of LLMs' abilities to generate words in the dialect of Mainz, demonstrating that existing models fail to produce correct dialectal words.\footnote{See Appendix~\ref{ap:code} for code and dataset resources.}

\section{Related Work}

\paragraph{German Dialect Datasets} Prior work has developed parallel dialect–Standard German dictionaries \cite[\textit{inter alia}]{haddow-etal-2013-corpus, artemova-plank-2023-low, litschko2025makelettercountbuilding}, but none address the Mainz dialect. \citet{blaschke-etal-2023-survey} compile over 80 German dialect corpora, including Rhine-Franconian varieties, however these do not cover Meenzerisch and consist only of unstructured text corpora.

\paragraph{NLP for German Dialects} A recent survey by \citet{blaschke-etal-2023-survey} shows that many German dialect speakers wish to use LLMs in their own dialects, highlighting the need for dialect-aware systems. Yet, studies reveal strong performance gaps between standard and dialectal German varieties across tasks \cite{kantharuban-etal-2023-quantifying, peng-etal-2024-sebastian, Blaschke_2025, bar-etal-2025-swiss, munoz-ortiz-etal-2025-evaluating}. Efforts to reduce this gap include noise injection during fine-tuning \cite{peng-etal-2024-sebastian} and pixel-based modeling that treats text as images \cite{munoz-ortiz-etal-2025-evaluating}. Recent work even demonstrates that LLMs systematically discriminate speakers of German dialects compared to Standard German speakers, indicating significant bias \cite{bui2025largelanguagemodelsdiscriminate}. 
Overall, however, Meenzerisch has so far remained unexplored due to the lack of suitable digital resources.

\section{Dataset: A Dictionary for the Dialect of Mainz}

\subsection{``Meenzerisch''---A German Dialect}
\begin{figure*}
    \centering
    \includegraphics[width=0.85\linewidth]{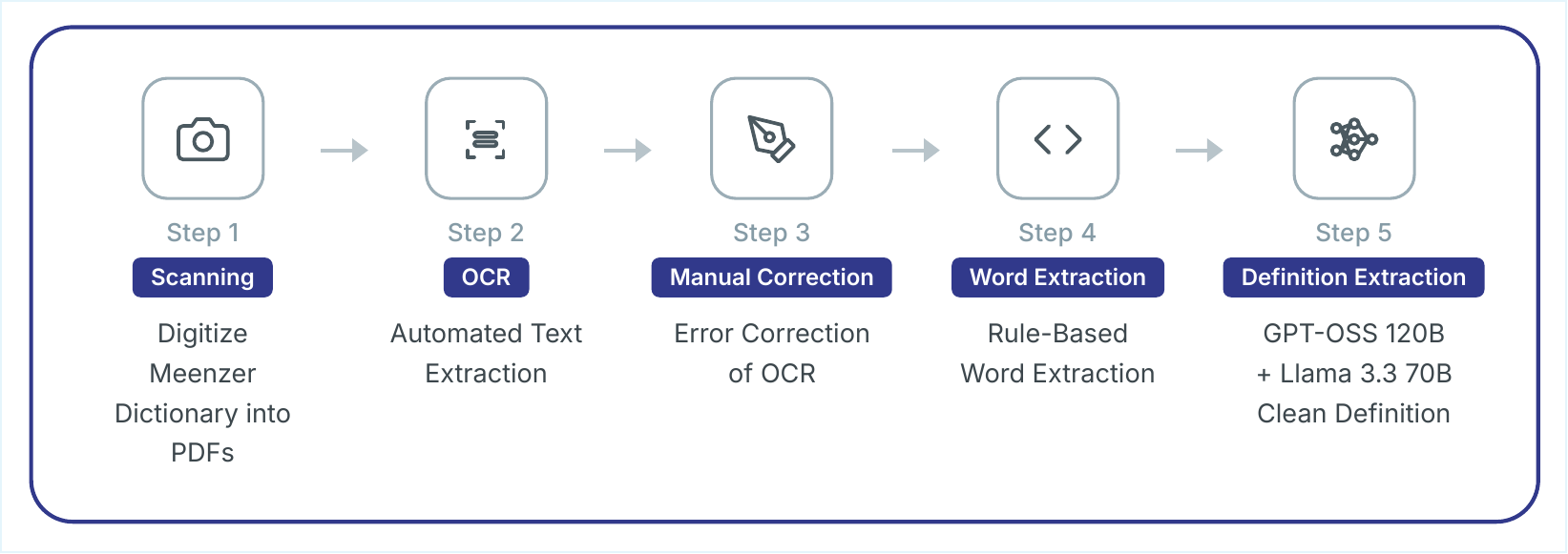}
    \caption{\textbf{Dataset Creation Pipeline.} Overview of the semi-automatic five-step process used to create the Mainz Dialect Dataset.}    \label{fig:pipeline}
\end{figure*}

\paragraph{Linguistic Classification and Distribution} The dialect of the city of Mainz (name of the city in dialect: \textit{Meenz}; also: \textit{M\"{a}\"{a}nz}) in Germany (name of the dialect in Standard German: \textit{Mainzerisch}, in dialect: \textit{Meenzerisch}; also \textit{Määnzerisch}) is a subvariety of the Rhine-Franconian dialect group of the German language. It is not only spoken in Mainz, but---in several local variations---in the region of Rhine Hesse (also: \textit{Rhenish Hesse}; Standard German: \textit{Rheinhessen}, in dialect: \textit{Rhoihesse}), which comprises the cities of Mainz, Bingen, Alzey and Worms (name of the regional dialect: \textit{Rheinhessisch}, in dialect: \textit{Rhoihessisch}).

Typically, native speakers of Meenzerisch do not distinguish between the phonemes “ch” [ç] 
and “sch” 
[\textesh]---everything is pronounced as “sch” 
[\textesh]. 
As Standard German does make the distinction, they usually learn at primary school that there is a difference between these phonemes. In Meenzerisch, ``wir'' (\textit{we}) and ``mir'' (\textit{me}) are both pronounced as ``mir'' or ``mer'', resulting in identical forms---also a characteristic causing confusion to Meenzerisch speaking children when faced with Standard German as they start kindergarten or primary school.

\paragraph{Historical Influences} During recent centuries, Meenzerisch has been strongly influenced by French. Among other historical events not mentioned here, Napoleon Bonaparte made Mainz (in French: \textit{Mayence}) a capital of the department ``Mont-Tonnerre'' of the French Empire. Typically, in Meenzerisch, words taken over from French are stressed on the first syllable but not on the last as in standard French \cite{mainz01}.

Furthermore, Mainz was one of the three important Jewish communities in medieval Germany (SchUM communities with Speyer (Yiddish: \textit{Spira}), Worms (Yiddish: \textit{Warmaisa}) and Mainz (Yiddish: \textit{Magenza}) \cite{levi1927magenza}. Thus, over a long period of time, Yiddish and Meenzerisch also influenced each other \cite{mainz01}.

Over the last decades, the number of native speakers has been diminishing. Simultaneously, Meenzerisch has been losing unique words and idioms. Neologisms, a typical feature of a living and evolving language, can hardly be observed today \cite{mainz01}. 
Meenzerisch is slowly being absorbed into a so-called regiolect of the Rhine-Main area \cite{vorberger2019regionalsprache}. 

\paragraph{Sociolinguistic Situation and Cultural Role} The slow attrition of Meenzerisch is due to a variety of factors which are common causes for the death of many low-resource languages and dialects (see, e.g., \citet{msila2011mama} for the situation of isiXhosa).
Another factor is the preference for Standard German among parents and educators, driven by the negative stereotypes dialects carry regarding educational status \cite{niemann1964landwirtschaft, EichingerGartigPlewnia2014, trillhasse}. Furthermore, the widespread presence of mass media and the internet has gradually introduced English and Standard German influences into regional speech. Additionally, social mobility contributes to dialect shift: in linguistically mixed households, the dialect is less likely to be transmitted to the next generation.

Nevertheless, Meenzerisch is still strongly anchored in the local culture of the city of Mainz. It can be characterised as the semi-official language used in the carnival of Mainz (Standard German: \textit{Mainzer Fastnacht}, in dialect: \textit{Meenzer Fassenacht}), a very popular festival starting annually on November 11 and ending on Ash Wednesday. Furthermore, Meenzerisch is the colloquial language in contact with local authorities, in the city center of Mainz, and also in the football stadium of Mainz and by supporters of the local football team.

Meenzerisch is an essential cornerstone of local culture, tradition and lifestyle and is, thus, according to many speakers deserving of preservation efforts:
``Ei natürlich! Du glaabst doch nit im Ernst, dass unser Muddersprooch verschwindt! Sowas is Kuldur! Verstehste! Und Kuldur ist wischtisch!'' \cite{mainz01}.\footnote{English translation: ``Of course! You don't really believe that our native language will die out, do you? It's culture! You know? And culture is important!''}

\subsection{Dataset Creation} \label{sec:dataset}

We create a dataset consisting of words in Meenzerisch together with a description of their meaning in Standard German by extracting definitions from a physical copy of the “Mainzer Wörterbuch” (English: \textit{dictionary of the dialect of Mainz}) by Karl Schramm \cite{schramm1966mainzer}. This is done semi-automatically by performing the following five steps: 1) scanning the book; 2) optical character recognition (OCR); 3) manual correction; 4) extraction of the word or expression from each line with a rule-based Python script; and 5) using an LLM to extract the definition from each line. Step 5 is necessary since many entries in the original book contain additional information, which would make it tricky to automatically evaluate an LLM’s understanding or generation skills, e.g., example sentences or words similar to the explained one. We ignore all such additional information for the purpose of this research, but highlight that it is valuable knowledge for future work, for instance when adapting LLMs to the dialect of Mainz.
We show an overview of our pipeline in Figure \ref{fig:pipeline}.

\paragraph{Scanning and OCR} We use a commercial app to create PDFs from pictures taken of the pages of the ``Mainzer W\"{o}rterbuch''. We use the same app for OCR to convert the PDFs into text files.

\paragraph{Manual Clean-Up of OCR Output}
While a manual inspection of the OCR output reveals that the quality is generally high, a common mistake is missing line breaks. They are problematic as, in the original book, new words are recognizable by starting a new line. Thus, in order to increase the quality of the subsequent automatic word and definition extraction, we manually add in missing line breaks to make sure each line contains exactly one word, together with its definition. While doing to, we also correct other obvious mistakes. This manual clean-up required approximately four hours.

\paragraph{Automatic Extraction using LLMs}
To isolate the definition of each word, excluding explanations and examples, we employ an LLM for automated extraction. Specifically, we use GPT-OSS 120B \cite{openai2025gptoss120bgptoss20bmodel} with a medium reasoning effort. The model is instructed to preserve the original wording and to number multiple meanings when present. We provide six handpicked, manually extracted examples as demonstrations within the prompt (see prompt in Appendix \ref{ap:prompt_dataset}).

Furthermore, we clean each extracted definition from OCR-induced noise, e.g., unnecessary hyphens or special characters, using Llama-3.3 70B \cite{grattafiori2024llama3herdmodels} to automatically normalize and sanitize the text, see Appendix \ref{ap:prompt_dataset} for the full prompt.

\paragraph{Quality Control} To evaluate the quality of our extraction procedure, one of the authors, who is a native speaker of Standard German, manually assesses a random subsample of 100 entries by comparing the extracted model outputs to the original book's Standard German definitions. We distinguish between major and minor errors. Major errors occur when the extracted meaning does not correspond to the intended definition, whereas minor errors refer to cases where redundant or irrelevant information is included, e.g., the additional tag ``Bildwort'' (\textit{English: figurative word}). Our evaluation reveals that 12\% of the entries contain minor errors, while only 6\% exhibit major errors. Given the low rate of major errors, we consider the overall extraction quality satisfactory.

\begin{table}[t]
\centering
\small
\begin{tabular}{lc}
\toprule
\textbf{Category} & \textbf{Value} \\
\midrule
\multicolumn{2}{l}{\textit{Dataset Size}} \\
\quad Total extracted pairs & 2,471 \\
\quad Extracted pairs after clean-up & 2,351 \\
\midrule
\multicolumn{2}{l}{\textit{Word Length}} \\
\quad Median length (characters) & 8 \\
\midrule
\multicolumn{2}{l}{\textit{Definition Length}} \\
\quad Avg. definitions per word & 1.37 \\
\quad Median length (characters) & 31 \\
\quad Median length (words) & 4 \\
\midrule
\multicolumn{2}{l}{\textit{Dataset Split}} \\
\quad Training set & 250 \\
\quad Development set & 250 \\
\quad Test set & 1,851 \\
\bottomrule
\end{tabular}
\caption{\textbf{Overview of the Mainz Dialect Dataset.} Summary of the extracted and cleaned dataset.}
\label{tab:dataset_overview}
\end{table}

\paragraph{Copyright Considerations}  We create our corpus according to §60d on Text and Data Mining of the German Urheberrechts-Wissensgesellschafts-Gesetz (English: \textit{law on copyright in the knowledge economy}). This law allows text and data mining with the goal to create corpora for non-commercial scientific research. It further allows to make the resulting corpora accessible to a defined group of people for joint scientific research or for control of the quality of scientific research. Thus, we will make our dataset available to other researchers upon request under the CC BY-NC-ND 4.0 license.

\subsection{Dataset Statistics}

We report a summary of the dataset in Table \ref{tab:dataset_overview}.

\paragraph{Number of Samples} After the extraction process, we obtain a total of 2,471 word–definition pairs. We remove samples containing missing or invalid definitions, resulting in a cleaned dataset of 2,351 valid entries. Table~\ref{tab:examples} shows representative examples from the dataset.

\paragraph{Word and Definition Length}
The median length of dialect words is 8 characters. Since some definitions include multiple meanings for a single word, we first report that each word contains an average of 1.37 meanings. Meanings have a median length of 31 characters and 4 words.

\paragraph{Dataset Split} The dataset is divided into 250 training, 250 development, and 1,851 test samples. We intentionally provide a larger test set than training data, as the main goal of this benchmark is thorough evaluation rather than model training.

\begin{table}[t]
    \centering
    \small
    \setlength{\tabcolsep}{6pt}
    \begin{tabular}{
        >{\RaggedRight\arraybackslash}p{2.4cm}  
        >{\RaggedRight\arraybackslash}p{4.5cm}  
    }
        \toprule
        \textbf{Word in Meenzerisch} & \textbf{Standard German Definition} \\ \midrule
        \textit{Aaweiderworschd} & Salzgurke (English: \textit{pickled cucumber}) \\
        \textit{Bitzelwasser} & Mineralwasser mit Kohlensäuregehalt (English:  \textit{Carbonated mineral water}) \\
        \textit{Mauldabbezierer} & scherzhaftes Lob für einen guten Wein [...]  (\textit{English: playful compliment for a good wine}) \\
        \textit{Rachebutzer} & saurer Wein (English: \textit{sour wine}) \\
        \textit{Schlobb} & 1. Schleife; 2. Knoten; 3. [...] (English: \textit{1. bow; 2. knot}) \\
        \textit{Schimmes} & Hunger (English: \textit{hunger})  \\
        \textit{Schwollescheer} & Angehörigen einer Reitertruppe (English: \textit{members of a cavalry troop}) \\
        \bottomrule
    \end{tabular}
    \caption{\textbf{Examples from Our Dataset.} Samples illustrating the mapping between dialect words and their corresponding Standard German definitions.}
    \label{tab:examples}
\end{table}

\begin{table*}[t]
    \centering
    \small\setlength{\tabcolsep}{15pt}
    \begin{tabular}{l|cc|c}
        \toprule
        \textbf{Model} & \multicolumn{2}{c|}{\textbf{Dialect of Mainz}} & \textbf{English} \\ 
        & \textbf{Dev Set} & \textbf{Test Set} & \textbf{Test Set} \\ 
        \midrule
        \multicolumn{4}{c}{\textit{Large Models}} \\ 
        \midrule
        GPT-OSS 120B (Thinking) & 5.60\% & 4.92\% & 91.19\% \\
        Llama-3.3 70B & 8.80\% & 6.27\% & 91.69\% \\
        Leo-HessianAI 70B & 4.40\% & 4.81\% & 81.88\% \\
        Qwen-2.5 72B & 4.80\% & 5.35\% & 90.19\% \\
        \midrule
        \multicolumn{4}{c}{\textit{Medium Models}} \\ 
        \midrule
        Aya Expanse 32B & 3.60\% & 4.97\% & 89.89\% \\
        Qwen-3 30B & 5.60\% & 3.84\% & 90.39\% \\
        Gemma-3 27B & 5.60\% & 5.51\% & 90.49\% \\ 
        Phi-4 14B & 9.60\% & 5.40\% & 90.59\% \\
        \midrule
        \multicolumn{4}{c}{\textit{Small Models}} \\ 
        \midrule
        Llama-3.1 8B & 3.60\% & 2.55\% & 85.29\% \\
        Qwen-2.5 7B & 4.40\% & 3.08\% & 83.78\% \\
        Gemma-3 4B & 2.80\% & 2.38\% & 80.08\% \\ 
        Qwen-3 4B & 1.60\% & 1.84\% & 77.18\% \\ 
        \midrule
        \textbf{Average} & \textbf{5.03\%} & \textbf{4.24\%} & \textbf{86.89\%} \\ 
        \bottomrule
    \end{tabular}
    \caption{\textbf{Accuracy of Definition Generation.} Development and test set accuracy of LLMs for the Mainz dialect, and test set accuracy for the English baseline.}
    \label{tab:meaning_accuracy}
\end{table*}

\section{Large Language Models}

We benchmark a diverse set of open-source, instruction-tuned models covering both dense and mixture-of-experts architectures, thinking and non-thinking models, of varying scale and multilingual coverage. Some models are primarily pretrained on human-generated data, while others rely heavily on LLM-generated synthetic data. All predictions are generated using greedy decoding.

\paragraph{Llama Models.} We evaluate two models from the Llama family: Llama-3.1~8B and Llama-3.3~70B by \citet{grattafiori2024llama3herdmodels}. Both are dense, decoder-only transformer architectures.

\paragraph{GPT-OSS-120B.} GPT-OSS~120B~(Thinking) by \citet{openai2025gptoss120bgptoss20bmodel} is a mixture-of-experts model designed for explicit thinking.

\paragraph{Qwen Models.} We include Qwen~3~4B~(Instruct) and Qwen~2.5~(7B,~72B)~\cite{qwen3technicalreport, qwen2025qwen25technicalreport}. Both are dense, decoder-only transformer models. Furthermore, we include Qwen3 30B (Instruct), which is a mixture-of-experts model. 

\paragraph{Gemma Models.} The Gemma~3~models (4B,~27B)~\cite{gemmateam2025gemma3technicalreport} are dense, decoder-only transformers optimized for multilingual and multimodal tasks, supporting over 140 languages. 

\paragraph{Leo-HessianAI Model.} Leo-HessianAI 70B \cite{pluester2023leolm} is a dense transformer model specifically optimized for German-language tasks. 

\paragraph{Phi~4.} Phi~4 (14B) \cite{abdin2024phi4technicalreport} is a compact, dense, decoder-only model emphasizing training on high-quality data. This model is pretrained on a large quantity of synthetic data.

\paragraph{Aya-Expanse Model.} Aya~Expanse~32B~\cite{dang2024ayaexpansecombiningresearch} is a multilingual, dense, decoder-only model trained on high-quality multilingual data.

\section{Experiment 1: Definition Generation} \label{sec:understanding}
In this experiment, we investigate the understanding capabilities of LLMs, defined as their ability to accurately infer and convey the meanings of Meenzerisch words in Standard German.

\subsection{Prompt Design}

We instruct the LLMs to generate one valid definition of each given word, assuming that the word is used in the dialect of Mainz and that the definition should reflect its meaning within this dialectal context. See Appendix \ref{ap:prompts_1} for the full prompt.

\subsection{Evaluation} \label{sec:eval_understanding}

\paragraph{Automatic Evaluation}
We employ an LLM-as-a-judge (\textsc{LLMaaJ}) approach for automatic evaluation, using Llama 3.3 70B as the judge. The evaluation is binary, labeling each pair of definitions as either equal or unequal. Since some words in the dictionary contain multiple reference definitions, we consider a generated definition correct if it matches at least one of the ground-truth definitions. We report the full prompt in Appendix \ref{ap:prompts_1}.

\paragraph{Manual Verification}
To evaluate the reliability of our \textsc{LLMaaJ} approach, we manually inspect a random subset of 100 samples, 50 labeled as unequal and 50 as equal by the model. We find that, for the unequal cases, the model's judgments are fully accurate (100\%). For the examples labeled as equal, we observe an accuracy of 92\%. These results indicate that our automated judgments yield accuracy values that approximate the upper limit of performance. We therefore consider this level of agreement sufficient for the purposes of our study.

\paragraph{English Comparison}
To establish a capacity baseline, we evaluate all LLMs on an English dictionary dataset \cite{anthony_therrien_2024}, sampling 2,000 entries from 42,052. Because English is heavily represented in pretraining data, this approximates the models' best-case performance on dictionary definition generation. We apply the identical inference and evaluation pipeline as in the dialect setting, differing only in prompting the models to produce English definitions to match the reference.

\begin{table*}[t]
    \centering
    \small \setlength{\tabcolsep}{15pt}
    \begin{tabular}{l|ccc}
        \toprule
        \textbf{Model} & \multicolumn{2}{c}{\textbf{Dialect of Mainz}} & \textbf{English} \\ 
        & \textbf{Dev Set} & \textbf{Test Set} & \textbf{Test Set} \\ 
        \midrule
        \multicolumn{4}{c}{\textit{Large Models}} \\ 
        \midrule
        GPT-OSS 120B (Thinking) & 2.40\% & 1.51\% & 82.50\% \\
        Llama-3.3 70B & 0.80\% & 1.03\% & 69.00\% \\
        Leo-HessianAI 70B & 0.80\% & 0.92\% & 47.10\% \\
        Qwen-2.5 72B & 0.80\% & 0.43\% & 66.90\% \\ 
        \midrule
        \multicolumn{4}{c}{\textit{Medium Models}} \\ 
        \midrule
        Aya Expanse 32B & 1.20\% & 0.59\% & 59.50\% \\
        Qwen-3 30B & 0.40\% & 0.27\% & 60.10\% \\
        Gemma-3 27B & 1.20\% & 0.43\% & 65.60\% \\ 
        Phi-4 14B & 1.60\% & 0.49\% & 64.10\% \\ 
        \midrule
        \multicolumn{4}{c}{\textit{Small Models}} \\ 
        \midrule
        Llama-3.1 8B & 0.00\% & 0.54\% & 52.10\% \\ 
        Qwen-2.5 7B & 0.00\% & 0.16\% & 49.55\% \\
        Gemma-3 4B & 0.00\% & 0.16\% & 43.24\% \\ 
        Qwen-3 4B & 0.00\% & 0.16\% & 45.50\% \\
        \midrule
        \textbf{Average} & \textbf{0.77\%} & \textbf{0.56\%} & \textbf{58.77\%} \\ 
        \bottomrule
    \end{tabular}
    \caption{\textbf{Accuracy of Word Generation given Explanation.} Development and test set accuracy of LLMs in generating the correct word for the Mainz dialect, and test set accuracy for the English baseline.}
    \label{tab:generation_accuracy}
\end{table*}

\subsection{Results and Discussion}

We report the test set results for Experiment 1 in Table \ref{tab:meaning_accuracy}. We additionally include the development set results to enable future comparisons.

Overall, we see low accuracy for all models: the average is just 4.24\%, with the best performing model, Llama-3.3 70B, reaching 6.27\% and the worst model, Qwen-3 4B, reaching only 1.84\%. 

Across model sizes, differences are modest, given the generally low accuracy levels. However, small models show the worst performance, with the best small model, Qwen-2.5 7B, having a lower accuracy than the worst medium model, Qwen-3 30B. Medium and large model performances overlap: the top three medium models each outperform at least one large model. However, the overall best model, Llama-3.3 70B, is a large model.

Comparing to English (rightmost column of Table \ref{tab:meaning_accuracy}), we see that, in contrast, the same models achieve an average accuracy of 86.89\% when generating English definitions of English words. Even the worst model, Qwen-3 4B, reaches an accuracy of 77.18\%. This shows that our LLMs work correctly and are, in fact, strong models, but do not accurately capture the meanings of dialect words.

To sum up, Experiment 1 shows that \textbf{LLMs of all sizes struggle with understanding words in the dialect of Mainz}.

\section{Experiment 2: Dialect Word Generation}
\label{sec:generation}
We now investigate whether LLMs correctly generate a dialect word when provided with its meaning.

\subsection{Prompt Design}
Each model is prompted with a Standard German explanation and instructed to produce the equivalent word in the dialect of Mainz. The prompt explicitly requests a single, concise dialect term without additional commentary or reformulation. See Appendix \ref{ap:prompts_2} for the full prompt.

\subsection{Evaluation}

\paragraph{Automatic Evaluation} To assess accuracy, we perform a direct string comparison between the (cleaned) predicted and the gold-standard dialect word. This straightforward equality-based evaluation quantifies how accurately each model reproduces the correct dialect form, and we report the resulting accuracy for each model individually.

\paragraph{English Comparison}
We again compare to the performance on the English dictionary from Section \ref{sec:eval_understanding}. Using the same inference pipeline, we prompt the models to generate the corresponding English word for a given definition.

\subsection{Results and Discussion}

We report the results of Experiment 2 on the test set in Table \ref{tab:generation_accuracy}. We also add development set results.

The results for Experiment 2 are even worse than those for Experiment 1: the average accuracy is only 0.56\%. In fact, the accuracy of most models is below 1\%---the only exceptions are Llama-3.3 70B with 1.03\% and GPT-OSS 120B (Thinking) with 1.51\%. As for Experiment 1, we see that small models, on average, perform worse than medium models, which, in turn, are outperformed by large models on average. However, given the low accuracies overall, absolute differences are minimal.

Again, we also compare to the same task in English, i.e., generating English words, given their definitions. We also see lower accuracies than for Experiment 1, hinting at the fact that the task is more difficult. However, performances for English are much better than for Meenzerisch, with an average accuracy of 58.77\% and the worst model, Gemma-3 4B, still obtaining 43.24\% accuracy.

We conclude that \textbf{LLMs are largely unable to produce words in the dialect of Mainz, underscoring a gap in their generative capabilities.}

\section{Additional Experiments}

Given the low performance of all LLMs in our main experiments, we further conduct two smaller experiments, using only the best LLM, Llama-3.3 70B, to explore whether few-shot in-context learning and automatic rule extraction improve the performance of definition and word generation (see Sections~\ref{sec:understanding} and Section~\ref{sec:generation}).

\subsection{Few-Shot Learning}
We first investigate if few-shot in-context learning improves the generation of definitions. This approach is particularly promising for the task, as many dialect words follow systematic transformations from their Standard German counterparts. For example, in Meenzerisch, infinitives end on ``-ele'' as opposed to on ``-en'' in Standard German.

We provide the LLM with $k$ training examples in the prompt.
We tune $k$ on the development set by testing 1, 5, 10, 25, and 50 examples (see Table~\ref{tab:fewshot_dev}). We do not apply prompt tuning, as it would require extensive computational resources.
To decide which examples to include in the prompt, we compare two retrieval strategies: random selection and edit distance-based selection. 

\paragraph{Selection: Random}
Examples are randomly sampled from the training set.
For each test instance, a new random subset of examples is drawn.

\paragraph{Selection: Edit Distance-Based}
We select examples based on textual similarity to the target instance, measured by Levenshtein edit distance.
This approach prioritizes examples whose surface forms are similar to the test word, aiming to provide the model with relevant cues for generating accurate definitions.  We use edit distance-based selection only for definition generation, as we do not expect edit distance to be meaningful for the relatively longer descriptions---the input of the word generation task.

\begin{table}[t]
    \centering
    \small
    \begin{tabular}{lcc}
        \toprule
        \textbf{Shots} & \textbf{Definition} & \textbf{Word} \\
        & \textbf{Generation} & \textbf{Generation} \\
        \midrule
        \multicolumn{3}{c}{\textit{Random Selection}} \\         \midrule
        1  & 6.00\% & 1.60\% \\
        5  & 9.60\% & \textbf{2.00\%} \\
        10 & \textbf{11.20\%} & 1.60\% \\
        25 & 7.60\% & 1.20\% \\
        50 & 8.00\% & \textbf{2.00\%} \\
        \midrule
        \multicolumn{3}{c}{\textit{Edit Distance-Based Selection}} \\         \midrule
        1  & 8.00\% & -- \\
        5  & 10.00\% & -- \\
        10 & 9.20\% & -- \\
        25 & 9.20\% & --\\
        50 & \textbf{10.40\%} & -- \\
        \bottomrule
    \end{tabular}
    \caption{\textbf{Different Numbers of Few-shot Examples on the Development Set.} 
    Comparison of Llama-3.3~70B~Instruct across different numbers of in-context examples.}
    \label{tab:fewshot_dev}
\end{table}

\begin{table}[t]
    \centering
    \small
    \begin{tabular}{c|c}
        \toprule
        \textbf{Selection Method} & \textbf{Accuracy} \\ \midrule
        \multicolumn{2}{c}{\textbf{Definition Generation}} \\ \midrule
        \textit{Baseline: Zero-Shot} & \textit{6.27\%} \\ 
        Random Selection ($k=10$) & 7.67\% \\
        Edit Distance-Based Selection ($k=50$) & \textbf{9.24\%*} \\  \midrule   
        \multicolumn{2}{c}{\textbf{Word Generation}} \\ \midrule
        \textit{Baseline: Zero-Shot} & \textbf{\textit{1.03\%}} \\ 
        Random Selection ($k=5$) & \textbf{1.24\%} \\
        \bottomrule
    \end{tabular}
    \caption{\textbf{Few-shot Results for Llama-3.3 70B.} Asterisks (*) mark significant improvements (McNemar’s test, $p < .05$); bolding highlights the best significant score (or multiple scores where differences are not significant).}
    \label{tab:few-shot}
\end{table}

\paragraph{Results}
Table~\ref{tab:few-shot} summarizes our results on the test set and compares them to Llama-3.3 70B's performance during Experiment 1 and 2 as a baseline. 
We see that all in-context learning approaches improve upon the zero-shot results. However, the magnitude of this improvement varies: 
For definition generation, random example selection yields a modest improvement of 1.4\% points, which is not statistically significant. In contrast, edit distance-based selection results in a significant gain of 2.97\%. For word generation and random selection, performance improves by only 0.21\%, which is not significant. Overall, we find that few-shot in-context learning provides small benefits, especially with edit distance-based selection, although the absolute accuracy levels remain very low.

\subsection{Automatic Rule Generation} \label{sec:automatic_rule_generation}

\begin{figure}[t]
    \centering
    \begin{promptbox}[Dialect-to-Standard Mapping Rules]
        \ttfamily\small\raggedright
        \#\# Practical Application Rules\\[6pt]

        \#\#\# Step-by-Step Mapping Process\\[4pt]

        1. Identify Core Root\\
        - Extract the main semantic element from the dialect word\\
        - Example: `Klebberschulde` → focus on `Klebb` and `Schuld`\\[4pt]

        2. Apply Sound Transformations\\
        - Apply systematic sound changes based on patterns above\\
        - Example: `Klebb` → `Kleb` (consonant simplification)\\
    \end{promptbox}
    \caption{\textbf{Part of our File with Automatically Extracted Rules.} The final rules are fed into Llama-3.3 70B for definition and word generation.}
   \label{fig:rules}
\end{figure}

We further experiment with automatically extracting linguistic transformation rules from the training dataset and feeding them to an LLM---again Llama-3.3 70B---for definition and word generation. To this end, we input the entire training split into DeepSeek-R1~\cite{Guo2025} and prompt the model to identify systematic mappings between dialect words and their corresponding Standard German forms. We provide the complete prompt and a detailed description of the procedure in Section~\ref{sec:automatic_rule}. We subsequently use the model-generated summary of these extracted rules. While such mappings may not apply universally, the model is instructed to detect recurring similarities and phonological or morphological patterns. The extracted rules are presented in a concise, human-interpretable format, designed to be both concrete and comprehensive. The beginning of our file with extracted rules is shown in Figure \ref{fig:rules}. We provide all extracted rules in Appendix \ref{sec:automatic_rule}.

These extracted rules are subsequently incorporated into the prompts at inference time to support rule-augmented generation.

\begin{table}[t]
    \centering
    \small
    \begin{tabular}{c|c}
        \toprule
        \textbf{Selection Method} & \textbf{Accuracy} \\ \midrule
        \multicolumn{2}{c}{\textbf{Definition Generation}} \\ \midrule
        \textit{Baseline: Zero-Shot} & \textit{6.27\%} \\
        Injected Extracted Rules & \textbf{8.37\%*} \\ \midrule
        \multicolumn{2}{c}{\textbf{Word Generation}} \\ \midrule
        \textit{Baseline: Zero-Shot} & \textbf{\textit{1.03\%}} \\
        Injected Extracted Rules & \textbf{0.76\%} \\ 
        \bottomrule
    \end{tabular}
    \caption{\textbf{Results for Automatically Extracted Rules Injected into Llama-3.3 70B.} Asterisks (*) mark significant improvements (McNemar’s test, $p < .05$); bolding highlights the best significant score (or multiple scores where differences are not significant).}
    \label{tab:rulebased}
\end{table}

\paragraph{Results}
Table~\ref{tab:rulebased} summarizes the results and, again, compares them to Llama-3.3 70B’s
performance during Experiments 1 and 2. Injecting the extracted rules into the model increases accuracy for definition generation to 8.37\%. This suggests that helpful rules are indeed automatically derived from the word--definition pairs in the training set. In contrast, performance on word generation slightly decreases, though the change is not statistically significant. Overall, accuracy remains low, indicating that the models still struggle with Meenzerisch, even with the provided rules.

\section{Conclusion}

With this work, we present the first NLP research on Meenzerisch, the dialect of the German city of Mainz. We first introduce a dataset of words in the dialect together with descriptions of their meaning in Standard German. 
We then evaluate the dialect knowledge of multiple LLMs by asking them to generate definitions for the words as well as words for given definitions.
Our evaluation reveals that state-of-the-art LLMs struggle with both tasks. While few-shot in-context learning and automatic rule extraction yield modest gains, overall performance remains low, underscoring the need for additional research to make LLMs work for Meenzerisch.

\section{Limitations}
We note that our work primarily establishes missing model abilities. Although we explore few-shot in-context learning and automatically extracted rules to improve performance, these are just two possible prompting techniques. Further prompt tuning as well as finetuning can potentially increase performance and should be investigated in future work. In order to enable LLMs to work well for Meenzerisch, additional language resources will need to be created and included into model training. Additionally, leveraging other dialect–standard language pairs could further enrich the dataset and support more robust generalization across dialectal variation. Crucially, future methods should not rely solely on increased compute, as this effectively requires underrepresented cultures to bear higher costs to achieve equivalent service \cite{bui-etal-2025-generalization}.

Finally, we note that our evaluation is limited to word-level dialect understanding rather than full sentences. However, as the first resource of its kind, our work already reveals failures at the word level, suggesting that this foundational gap must be addressed before moving to more complex linguistic structures.

\section{Ethics Statements} 

Our dataset contains a small number of potentially offensive or sensitive words. In our training split, we identify 5 such instances out of 250 samples (2\%). As these expressions are authentic components of the dialect and carry linguistic relevance, we choose to retain them in the dataset. 

In adherence to the standards for working with minority groups \cite{liu-etal-2022-always,mager-etal-2023-ethical}, we actively engage with community members, with one native speaker joining this work as an author.

\section{Acknowledgment}
This work was supported by the Carl Zeiss Foundation through the TOPML project, grant number P2021-02-014.

\section{Bibliographical References}\label{sec:reference}
\bibliographystyle{lrec2026-natbib}
\bibliography{lrec2026-example}

@misc{openai2025gptoss120bgptoss20bmodel,
      title={gpt-oss-120b & gpt-oss-20b Model Card}, 
      author={OpenAI},
      year={2025},
      eprint={2508.10925},
      archivePrefix={arXiv},
      primaryClass={cs.CL},
      url={https://arxiv.org/abs/2508.10925}, 
}

@misc{mainz01,
    title = {Die kleine Sprachgeschichte. {M}eenzerisch -- {O}der: {W}arum die da so anders sprechen},
    author = {Hans-Peter Betz},
    year = {2010},
    howpublished = {\url{https://assets.deutschlandfunk.de/FILE_ea77b362cd2567bdbe5a34935531af4e/original.pdf}},
    note = {Accessed: 2025-10-15}
}

@book{levi1927magenza,
  title={Magenza: ein Sammelheft {\"u}ber das j{\"u}dische Mainz im f{\"u}nfhundertsten Todesjahre des Mainzer Gelehrten Maharil},
  author={Levi, Sali},
  year={1927},
  publisher={Habrith-Verlag-Gesellschaft}
}

@inproceedings{bui-etal-2025-generalization,
    title = "On Generalization across Measurement Systems: {LLM}s Entail More Test-Time Compute for Underrepresented Cultures",
    author = "Bui, Minh Duc  and
      Park, Kyung Eun  and
      Glava{\v{s}}, Goran  and
      Schmidt, Fabian David  and
      Wense, Katharina Von Der",
    editor = "Che, Wanxiang  and
      Nabende, Joyce  and
      Shutova, Ekaterina  and
      Pilehvar, Mohammad Taher",
    booktitle = "Proceedings of the 63rd Annual Meeting of the Association for Computational Linguistics (Volume 1: Long Papers)",
    month = jul,
    year = "2025",
    address = "Vienna, Austria",
    publisher = "Association for Computational Linguistics",
    url = "https://aclanthology.org/2025.acl-long.1032/",
    doi = "10.18653/v1/2025.acl-long.1032",
    pages = "21262--21276",
    ISBN = "979-8-89176-251-0"
}

@article{msila2011mama,
  title={“Mama does not speak that (language) to me”: indigenous languages, educa-tional opportunity and black African preschoolers},
  author={Msila, Vuyisile},
  journal={South African Journal of Childhood Education},
  volume={1},
  number={1},
  pages={20},
  year={2011}
}

@book{schramm1966mainzer,
  title={Mainzer W{\"o}rterbuch},
  author={Schramm, Karl},
  publisher={Dr. Hanns Krach},
  year={1966}
}

@book{vorberger2019regionalsprache,
  title={Regionalsprache in Hessen: eine Untersuchung zu Sprachvariation und Sprachwandel im mittleren und s{\"u}dlichen Hessen},
  author={Vorberger, Lars},
  year={2019},
  publisher={Franz Steiner Verlag Stuttgart}
}

@misc{litschko2025makelettercountbuilding,
      title={Make Every Letter Count: Building Dialect Variation Dictionaries from Monolingual Corpora}, 
      author={Robert Litschko and Verena Blaschke and Diana Burkhardt and Barbara Plank and Diego Frassinelli},
      year={2025},
      eprint={2509.17855},
      archivePrefix={arXiv},
      primaryClass={cs.CL},
      url={https://arxiv.org/abs/2509.17855}, 
}

@inproceedings{haddow-etal-2013-corpus,
    title = "Corpus development for machine translation between standard and dialectal varieties",
    author = "Haddow, Barry  and
      Hern{\'a}ndez, Adolfo  and
      Neubarth, Friedrich  and
      Trost, Harald",
    editor = "Vertan, Cristina  and
      Slavcheva, Milena  and
      Osenova, Petya",
    booktitle = "Proceedings of the Workshop on Adaptation of Language Resources and Tools for Closely Related Languages and Language Variants",
    month = sep,
    year = "2013",
    address = "Hissar, Bulgaria",
    publisher = "INCOMA Ltd. Shoumen, BULGARIA",
    url = "https://aclanthology.org/W13-5303/",
    pages = "7--14"
}

@inproceedings{blaschke-etal-2023-survey,
    title = "A Survey of Corpora for {G}ermanic Low-Resource Languages and Dialects",
    author = "Blaschke, Verena  and
      Schuetze, Hinrich  and
      Plank, Barbara",
    editor = {Alum{\"a}e, Tanel  and
      Fishel, Mark},
    booktitle = "Proceedings of the 24th Nordic Conference on Computational Linguistics (NoDaLiDa)",
    month = may,
    year = "2023",
    address = "T{\'o}rshavn, Faroe Islands",
    publisher = "University of Tartu Library",
    url = "https://aclanthology.org/2023.nodalida-1.41/",
    pages = "392--414"
}

@inproceedings{bar-etal-2025-swiss,
    title = "{S}wiss {G}erman Speech Translation and the Curse of Multidialectality",
    author = {B{\"a}r, Martin  and
      DeMarco, Andrea  and
      Labaka, Gorka},
    editor = "Salesky, Elizabeth  and
      Federico, Marcello  and
      Anastasopoulos, Antonis",
    booktitle = "Proceedings of the 22nd International Conference on Spoken Language Translation (IWSLT 2025)",
    month = jul,
    year = "2025",
    address = "Vienna, Austria (in-person and online)",
    publisher = "Association for Computational Linguistics",
    url = "https://aclanthology.org/2025.iwslt-1.15/",
    doi = "10.18653/v1/2025.iwslt-1.15",
    pages = "165--179",
    ISBN = "979-8-89176-272-5"
}

@inproceedings{Blaschke_2025,
   title={A Multi-Dialectal Dataset for German Dialect ASR and Dialect-to-Standard Speech Translation},
   url={http://dx.doi.org/10.21437/Interspeech.2025-318},
   DOI={10.21437/interspeech.2025-318},
   booktitle={Interspeech 2025},
   publisher={ISCA},
   author={Blaschke, Verena and Winkler, Miriam and Förster, Constantin and Wenger-Glemser, Gabriele and Plank, Barbara},
   year={2025},
   month=aug, pages={913–917},
   collection={interspeech_2025} 
}

@inproceedings{kantharuban-etal-2023-quantifying,
    title = "Quantifying the Dialect Gap and its Correlates Across Languages",
    author = "Kantharuban, Anjali  and
      Vuli{\'c}, Ivan  and
      Korhonen, Anna",
    editor = "Bouamor, Houda  and
      Pino, Juan  and
      Bali, Kalika",
    booktitle = "Findings of the Association for Computational Linguistics: EMNLP 2023",
    month = dec,
    year = "2023",
    address = "Singapore",
    publisher = "Association for Computational Linguistics",
    url = "https://aclanthology.org/2023.findings-emnlp.481/",
    doi = "10.18653/v1/2023.findings-emnlp.481",
    pages = "7226--7245"
}

@misc{bui2025largelanguagemodelsdiscriminate,
      title={Large Language Models Discriminate Against Speakers of German Dialects}, 
      author={Minh Duc Bui and Carolin Holtermann and Valentin Hofmann and Anne Lauscher and Katharina von der Wense},
      year={2025},
      eprint={2509.13835},
      archivePrefix={arXiv},
      primaryClass={cs.CL},
      url={https://arxiv.org/abs/2509.13835}, 
}

@book{niemann1964landwirtschaft,
  title={Die Landwirtschaft Niedersachsens, 1914-1964},
  author={Niemann, A.},
  lccn={65073296},
  url={https://books.google.de/books?id=FOo0AQAAIAAJ},
  year={1964},
  publisher={Landbuch-Verlag}
}

@book{EichingerGartigPlewnia2014,
  author    = {Eichinger, Ludwig M. and G{\"a}rtig, Anne-Kathrin and Plewnia, Albrecht},
  title     = {Aktuelle Spracheinstellungen in Deutschland},
  isbn      = {978-3-937241-28-9},
  pages     = {63 S.},
  year      = {2014},
  subject      = {Einstellung},
  language  = {de},
  publisher = {Institut für Deutsche Sprache
und Universität Mannheim}
}

@book{trillhasse,
author = {Trillhaase, Kerstin},
year = {2021},
month = {08},
pages = {},
publisher = {Logos Verlag Berlin},
title = {Der Einfluss der deutschen Dialekte Obersächsisch und Mittelbairisch auf die Wahrnehmung der Persönlichkeit},
isbn = {978-3-8325-5313-5}
}

@inproceedings{munoz-ortiz-etal-2025-evaluating,
    title = "Evaluating Pixel Language Models on Non-Standardized Languages",
    author = "Mu{\~n}oz-Ortiz, Alberto  and
      Blaschke, Verena  and
      Plank, Barbara",
    editor = "Rambow, Owen  and
      Wanner, Leo  and
      Apidianaki, Marianna  and
      Al-Khalifa, Hend  and
      Eugenio, Barbara Di  and
      Schockaert, Steven",
    booktitle = "Proceedings of the 31st International Conference on Computational Linguistics",
    month = jan,
    year = "2025",
    address = "Abu Dhabi, UAE",
    publisher = "Association for Computational Linguistics",
    url = "https://aclanthology.org/2025.coling-main.427/",
    pages = "6412--6419"
}

@inproceedings{peng-etal-2024-sebastian,
    title = "{S}ebastian, {B}asti, {W}astl?! Recognizing Named Entities in {B}avarian Dialectal Data",
    author = "Peng, Siyao  and
      Sun, Zihang  and
      Shan, Huangyan  and
      Kolm, Marie  and
      Blaschke, Verena  and
      Artemova, Ekaterina  and
      Plank, Barbara",
    editor = "Calzolari, Nicoletta  and
      Kan, Min-Yen  and
      Hoste, Veronique  and
      Lenci, Alessandro  and
      Sakti, Sakriani  and
      Xue, Nianwen",
    booktitle = "Proceedings of the 2024 Joint International Conference on Computational Linguistics, Language Resources and Evaluation (LREC-COLING 2024)",
    month = may,
    year = "2024",
    address = "Torino, Italia",
    publisher = "ELRA and ICCL",
    url = "https://aclanthology.org/2024.lrec-main.1262/",
    pages = "14478--14493"
}

@inproceedings{artemova-plank-2023-low,
    title = "Low-resource Bilingual Dialect Lexicon Induction with Large Language Models",
    author = "Artemova, Ekaterina  and
      Plank, Barbara",
    editor = {Alum{\"a}e, Tanel  and
      Fishel, Mark},
    booktitle = "Proceedings of the 24th Nordic Conference on Computational Linguistics (NoDaLiDa)",
    month = may,
    year = "2023",
    address = "T{\'o}rshavn, Faroe Islands",
    publisher = "University of Tartu Library",
    url = "https://aclanthology.org/2023.nodalida-1.39/",
    pages = "371--385"
}

@article{Guo2025,
  author    = {Daya Guo and Dejian Yang and Haowei Zhang and Junxiao Song and Peiyi Wang and Qihao Zhu and Runxin Xu and Ruoyu Zhang and Shirong Ma and Xiao Bi and Xiaokang Zhang and Xingkai Yu and Yu Wu and Z. F. Wu and Zhibin Gou and Zhihong Shao and Zhuoshu Li and Ziyi Gao and Aixin Liu and Bing Xue and Bingxuan Wang and Bochao Wu and Bei Feng and Chengda Lu and Chenggang Zhao and Chengqi Deng and Chenyu Zhang and Chong Ruan and Damai Dai and Deli Chen and Dongjie Ji and Erhang Li and Fangyun Lin and Fucong Dai and Fuli Luo and Guangbo Hao and Guanting Chen and Guowei Li and Han Bao and Hanwei Xu and Haocheng Wang and Honghui Ding and Huajian Xin and Huazuo Gao and Hui Qu and Hui Li and Jianzhong Guo and Jiashi Li and Jiawei Wang and Jingchang Chen and Jingyang Yuan and Junjie Qiu and Junlong Li and J. L. Cai and Jiaqi Ni and Jian Liang and Jin Chen and Kai Dong and Kai Hu and Kaige Gao and Kang Guan and Kexin Huang and Kuai Yu and Lean Wang and Lecong Zhang and Liang Zhao and Litong Wang and Liyue Zhang and Lei Xu and Leyi Xia and Mingchuan Zhang and Minghua Zhang and Minghui Tang and Meng Li and Miaojun Wang and Mingming Li and Ning Tian and Panpan Huang and Peng Zhang and Qiancheng Wang and Qinyu Chen and Qiushi Du and Ruiqi Ge and Ruisong Zhang and Ruizhe Pan and Runji Wang and R. J. Chen and R. L. Jin and Ruyi Chen and Shanghao Lu and Shangyan Zhou and Shanhuang Chen and Shengfeng Ye and Shiyu Wang and Shuiping Yu and Shunfeng Zhou and Shuting Pan and S. S. Li and Shuang Zhou and Shaoqing Wu and Tao Yun and Tian Pei and Tianyu Sun and T. Wang and Wangding Zeng and Wanjia Zhao and Wen Liu and Wenfeng Liang and Wenjun Gao and Wenqin Yu and Wentao Zhang and W. L. Xiao and Wei An and Xiaodong Liu and Xiaohan Wang and Xiaokang Chen and Xiaotao Nie and Xin Cheng and Xin Liu and Xin Xie and Xingchao Liu and Xinyu Yang and Xinyuan Li and Xuecheng Su and Xuheng Lin and X. Q. Li and Xiangyue Jin and Xiaojin Shen and Xiaosha Chen and Xiaowen Sun and Xiaoxiang Wang and Xinnan Song and Xinyi Zhou and Xianzu Wang and Xinxia Shan and Y. K. Li and Y. Q. Wang and Y. X. Wei and Yang Zhang and Yanhong Xu and Yao Li and Yao Zhao and Yaofeng Sun and Yaohui Wang and Yi Yu and Yichao Zhang and Yifan Shi and Yiliang Xiong and Ying He and Yishi Piao and Yisong Wang and Yixuan Tan and Yiyang Ma and Yiyuan Liu and Yongqiang Guo and Yuan Ou and Yuduan Wang and Yue Gong and Yuheng Zou and Yujia He and Yunfan Xiong and Yuxiang Luo and Yuxiang You and Yuxuan Liu and Yuyang Zhou and Y. X. Zhu and Yanping Huang and Yaohui Li and Yi Zheng and Yuchen Zhu and Yunxian Ma and Ying Tang and Yukun Zha and Yuting Yan and Z. Z. Ren and Zehui Ren and Zhangli Sha and Zhe Fu and Zhean Xu and Zhenda Xie and Zhengyan Zhang and Zhewen Hao and Zhicheng Ma and Zhigang Yan and Zhiyu Wu and Zihui Gu and Zijia Zhu and Zijun Liu and Zilin Li and Ziwei Xie and Ziyang Song and Zizheng Pan and Zhen Huang and Zhipeng Xu and Zhongyu Zhang and Zhen Zhang and DeepSeek-AI},
  title     = {DeepSeek-R1 incentivizes reasoning in LLMs through reinforcement learning},
  journal   = {Nature},
  year      = {2025},
  volume    = {645},
  number    = {8081},
  pages     = {633--638},
  doi       = {10.1038/s41586-025-09422-z},
  url       = {https://doi.org/10.1038/s41586-025-09422-z},
  issn      = {1476-4687}
}

@misc{anthony_therrien_2024,
	title={Dictionary of English Words and Definitions},
	url={https://www.kaggle.com/dsv/9456760},
	DOI={10.34740/KAGGLE/DSV/9456760},
	publisher={Kaggle},
	author={Anthony Therrien},
	year={2024}
}

@misc{dang2024ayaexpansecombiningresearch,
      title={Aya Expanse: Combining Research Breakthroughs for a New Multilingual Frontier}, 
      author={John Dang and Shivalika Singh and Daniel D'souza and Arash Ahmadian and Alejandro Salamanca and Madeline Smith and Aidan Peppin and Sungjin Hong and Manoj Govindassamy and Terrence Zhao and Sandra Kublik and Meor Amer and Viraat Aryabumi and Jon Ander Campos and Yi-Chern Tan and Tom Kocmi and Florian Strub and Nathan Grinsztajn and Yannis Flet-Berliac and Acyr Locatelli and Hangyu Lin and Dwarak Talupuru and Bharat Venkitesh and David Cairuz and Bowen Yang and Tim Chung and Wei-Yin Ko and Sylvie Shang Shi and Amir Shukayev and Sammie Bae and Aleksandra Piktus and Roman Castagné and Felipe Cruz-Salinas and Eddie Kim and Lucas Crawhall-Stein and Adrien Morisot and Sudip Roy and Phil Blunsom and Ivan Zhang and Aidan Gomez and Nick Frosst and Marzieh Fadaee and Beyza Ermis and Ahmet Üstün and Sara Hooker},
      year={2024},
      eprint={2412.04261},
      archivePrefix={arXiv},
      primaryClass={cs.CL},
      url={https://arxiv.org/abs/2412.04261}, 
}

@misc{qwen3technicalreport,
      title={Qwen3 Technical Report}, 
      author={Qwen, Team},
      year={2025},
      eprint={2505.09388},
      archivePrefix={arXiv},
      primaryClass={cs.CL},
      url={https://arxiv.org/abs/2505.09388}, 
}

@misc{pluester2023leolm,
  title        = {LeoLM/leo-hessianai-70b},
  author       = {Pl{\"u}ster, Bj{\"o}rn and Schuhmann, Christoph},
  year         = {2023},
  howpublished = {\url{https://huggingface.co/LeoLM/leo-hessianai-70b}},
  note         = {German-English bilingual language model based on Llama-2 70B, continued pretraining on German corpora. Released under the Llama 2 Community License.}
}

@misc{abdin2024phi4technicalreport,
      title={Phi-4 Technical Report}, 
      author={Marah Abdin and Jyoti Aneja and Harkirat Behl and Sébastien Bubeck and Ronen Eldan and Suriya Gunasekar and Michael Harrison and Russell J. Hewett and Mojan Javaheripi and Piero Kauffmann and James R. Lee and Yin Tat Lee and Yuanzhi Li and Weishung Liu and Caio C. T. Mendes and Anh Nguyen and Eric Price and Gustavo de Rosa and Olli Saarikivi and Adil Salim and Shital Shah and Xin Wang and Rachel Ward and Yue Wu and Dingli Yu and Cyril Zhang and Yi Zhang},
      year={2024},
      eprint={2412.08905},
      archivePrefix={arXiv},
      primaryClass={cs.CL},
      url={https://arxiv.org/abs/2412.08905}, 
}

@misc{gemmateam2025gemma3technicalreport,
      title={Gemma 3 Technical Report}, 
      author={Team Gemma and Aishwarya Kamath and Johan Ferret and Shreya Pathak and Nino Vieillard and Ramona Merhej and Sarah Perrin and Tatiana Matejovicova and Alexandre Ramé and Morgane Rivière and Louis Rouillard and Thomas Mesnard and Geoffrey Cideron and Jean-bastien Grill and Sabela Ramos and Edouard Yvinec and Michelle Casbon and Etienne Pot and Ivo Penchev and Gaël Liu and Francesco Visin and Kathleen Kenealy and Lucas Beyer and Xiaohai Zhai and Anton Tsitsulin and Robert Busa-Fekete and Alex Feng and Noveen Sachdeva and Benjamin Coleman and Yi Gao and Basil Mustafa and Iain Barr and Emilio Parisotto and David Tian and Matan Eyal and Colin Cherry and Jan-Thorsten Peter and Danila Sinopalnikov and Surya Bhupatiraju and Rishabh Agarwal and Mehran Kazemi and Dan Malkin and Ravin Kumar and David Vilar and Idan Brusilovsky and Jiaming Luo and Andreas Steiner and Abe Friesen and Abhanshu Sharma and Abheesht Sharma and Adi Mayrav Gilady and Adrian Goedeckemeyer and Alaa Saade and Alex Feng and Alexander Kolesnikov and Alexei Bendebury and Alvin Abdagic and Amit Vadi and András György and André Susano Pinto and Anil Das and Ankur Bapna and Antoine Miech and Antoine Yang and Antonia Paterson and Ashish Shenoy and Ayan Chakrabarti and Bilal Piot and Bo Wu and Bobak Shahriari and Bryce Petrini and Charlie Chen and Charline Le Lan and Christopher A. Choquette-Choo and CJ Carey and Cormac Brick and Daniel Deutsch and Danielle Eisenbud and Dee Cattle and Derek Cheng and Dimitris Paparas and Divyashree Shivakumar Sreepathihalli and Doug Reid and Dustin Tran and Dustin Zelle and Eric Noland and Erwin Huizenga and Eugene Kharitonov and Frederick Liu and Gagik Amirkhanyan and Glenn Cameron and Hadi Hashemi and Hanna Klimczak-Plucińska and Harman Singh and Harsh Mehta and Harshal Tushar Lehri and Hussein Hazimeh and Ian Ballantyne and Idan Szpektor and Ivan Nardini and Jean Pouget-Abadie and Jetha Chan and Joe Stanton and John Wieting and Jonathan Lai and Jordi Orbay and Joseph Fernandez and Josh Newlan and Ju-yeong Ji and Jyotinder Singh and Kat Black and Kathy Yu and Kevin Hui and Kiran Vodrahalli and Klaus Greff and Linhai Qiu and Marcella Valentine and Marina Coelho and Marvin Ritter and Matt Hoffman and Matthew Watson and Mayank Chaturvedi and Michael Moynihan and Min Ma and Nabila Babar and Natasha Noy and Nathan Byrd and Nick Roy and Nikola Momchev and Nilay Chauhan and Noveen Sachdeva and Oskar Bunyan and Pankil Botarda and Paul Caron and Paul Kishan Rubenstein and Phil Culliton and Philipp Schmid and Pier Giuseppe Sessa and Pingmei Xu and Piotr Stanczyk and Pouya Tafti and Rakesh Shivanna and Renjie Wu and Renke Pan and Reza Rokni and Rob Willoughby and Rohith Vallu and Ryan Mullins and Sammy Jerome and Sara Smoot and Sertan Girgin and Shariq Iqbal and Shashir Reddy and Shruti Sheth and Siim Põder and Sijal Bhatnagar and Sindhu Raghuram Panyam and Sivan Eiger and Susan Zhang and Tianqi Liu and Trevor Yacovone and Tyler Liechty and Uday Kalra and Utku Evci and Vedant Misra and Vincent Roseberry and Vlad Feinberg and Vlad Kolesnikov and Woohyun Han and Woosuk Kwon and Xi Chen and Yinlam Chow and Yuvein Zhu and Zichuan Wei and Zoltan Egyed and Victor Cotruta and Minh Giang and Phoebe Kirk and Anand Rao and Kat Black and Nabila Babar and Jessica Lo and Erica Moreira and Luiz Gustavo Martins and Omar Sanseviero and Lucas Gonzalez and Zach Gleicher and Tris Warkentin and Vahab Mirrokni and Evan Senter and Eli Collins and Joelle Barral and Zoubin Ghahramani and Raia Hadsell and Yossi Matias and D. Sculley and Slav Petrov and Noah Fiedel and Noam Shazeer and Oriol Vinyals and Jeff Dean and Demis Hassabis and Koray Kavukcuoglu and Clement Farabet and Elena Buchatskaya and Jean-Baptiste Alayrac and Rohan Anil and Dmitry and Lepikhin and Sebastian Borgeaud and Olivier Bachem and Armand Joulin and Alek Andreev and Cassidy Hardin and Robert Dadashi and Léonard Hussenot},
      year={2025},
      eprint={2503.19786},
      archivePrefix={arXiv},
      primaryClass={cs.CL},
      url={https://arxiv.org/abs/2503.19786}, 
}

@misc{qwen2025qwen25technicalreport,
      title={Qwen2.5 Technical Report}, 
      author={Qwen, Team and An Yang and Baosong Yang and Beichen Zhang and Binyuan Hui and Bo Zheng and Bowen Yu and Chengyuan Li and Dayiheng Liu and Fei Huang and Haoran Wei and Huan Lin and Jian Yang and Jianhong Tu and Jianwei Zhang and Jianxin Yang and Jiaxi Yang and Jingren Zhou and Junyang Lin and Kai Dang and Keming Lu and Keqin Bao and Kexin Yang and Le Yu and Mei Li and Mingfeng Xue and Pei Zhang and Qin Zhu and Rui Men and Runji Lin and Tianhao Li and Tianyi Tang and Tingyu Xia and Xingzhang Ren and Xuancheng Ren and Yang Fan and Yang Su and Yichang Zhang and Yu Wan and Yuqiong Liu and Zeyu Cui and Zhenru Zhang and Zihan Qiu},
      year={2025},
      eprint={2412.15115},
      archivePrefix={arXiv},
      primaryClass={cs.CL},
      url={https://arxiv.org/abs/2412.15115}, 
}

@misc{grattafiori2024llama3herdmodels,
      title={The Llama 3 Herd of Models}, 
      author={Aaron Grattafiori and Abhimanyu Dubey and Abhinav Jauhri and Abhinav Pandey and Abhishek Kadian and Ahmad Al-Dahle and Aiesha Letman and Akhil Mathur and Alan Schelten and Alex Vaughan and Amy Yang and Angela Fan and Anirudh Goyal and Anthony Hartshorn and Aobo Yang and Archi Mitra and Archie Sravankumar and Artem Korenev and Arthur Hinsvark and Arun Rao and Aston Zhang and Aurelien Rodriguez and Austen Gregerson and Ava Spataru and Baptiste Roziere and Bethany Biron and Binh Tang and Bobbie Chern and Charlotte Caucheteux and Chaya Nayak and Chloe Bi and Chris Marra and Chris McConnell and Christian Keller and Christophe Touret and Chunyang Wu and Corinne Wong and Cristian Canton Ferrer and Cyrus Nikolaidis and Damien Allonsius and Daniel Song and Danielle Pintz and Danny Livshits and Danny Wyatt and David Esiobu and Dhruv Choudhary and Dhruv Mahajan and Diego Garcia-Olano and Diego Perino and Dieuwke Hupkes and Egor Lakomkin and Ehab AlBadawy and Elina Lobanova and Emily Dinan and Eric Michael Smith and Filip Radenovic and Francisco Guzmán and Frank Zhang and Gabriel Synnaeve and Gabrielle Lee and Georgia Lewis Anderson and Govind Thattai and Graeme Nail and Gregoire Mialon and Guan Pang and Guillem Cucurell and Hailey Nguyen and Hannah Korevaar and Hu Xu and Hugo Touvron and Iliyan Zarov and Imanol Arrieta Ibarra and Isabel Kloumann and Ishan Misra and Ivan Evtimov and Jack Zhang and Jade Copet and Jaewon Lee and Jan Geffert and Jana Vranes and Jason Park and Jay Mahadeokar and Jeet Shah and Jelmer van der Linde and Jennifer Billock and Jenny Hong and Jenya Lee and Jeremy Fu and Jianfeng Chi and Jianyu Huang and Jiawen Liu and Jie Wang and Jiecao Yu and Joanna Bitton and Joe Spisak and Jongsoo Park and Joseph Rocca and Joshua Johnstun and Joshua Saxe and Junteng Jia and Kalyan Vasuden Alwala and Karthik Prasad and Kartikeya Upasani and Kate Plawiak and Ke Li and Kenneth Heafield and Kevin Stone and Khalid El-Arini and Krithika Iyer and Kshitiz Malik and Kuenley Chiu and Kunal Bhalla and Kushal Lakhotia and Lauren Rantala-Yeary and Laurens van der Maaten and Lawrence Chen and Liang Tan and Liz Jenkins and Louis Martin and Lovish Madaan and Lubo Malo and Lukas Blecher and Lukas Landzaat and Luke de Oliveira and Madeline Muzzi and Mahesh Pasupuleti and Mannat Singh and Manohar Paluri and Marcin Kardas and Maria Tsimpoukelli and Mathew Oldham and Mathieu Rita and Maya Pavlova and Melanie Kambadur and Mike Lewis and Min Si and Mitesh Kumar Singh and Mona Hassan and Naman Goyal and Narjes Torabi and Nikolay Bashlykov and Nikolay Bogoychev and Niladri Chatterji and Ning Zhang and Olivier Duchenne and Onur Çelebi and Patrick Alrassy and Pengchuan Zhang and Pengwei Li and Petar Vasic and Peter Weng and Prajjwal Bhargava and Pratik Dubal and Praveen Krishnan and Punit Singh Koura and Puxin Xu and Qing He and Qingxiao Dong and Ragavan Srinivasan and Raj Ganapathy and Ramon Calderer and Ricardo Silveira Cabral and Robert Stojnic and Roberta Raileanu and Rohan Maheswari and Rohit Girdhar and Rohit Patel and Romain Sauvestre and Ronnie Polidoro and Roshan Sumbaly and Ross Taylor and Ruan Silva and Rui Hou and Rui Wang and Saghar Hosseini and Sahana Chennabasappa and Sanjay Singh and Sean Bell and Seohyun Sonia Kim and Sergey Edunov and Shaoliang Nie and Sharan Narang and Sharath Raparthy and Sheng Shen and Shengye Wan and Shruti Bhosale and Shun Zhang and Simon Vandenhende and Soumya Batra and Spencer Whitman and Sten Sootla and Stephane Collot and Suchin Gururangan and Sydney Borodinsky and Tamar Herman and Tara Fowler and Tarek Sheasha and Thomas Georgiou and Thomas Scialom and Tobias Speckbacher and Todor Mihaylov and Tong Xiao and Ujjwal Karn and Vedanuj Goswami and Vibhor Gupta and Vignesh Ramanathan and Viktor Kerkez and Vincent Gonguet and Virginie Do and Vish Vogeti and Vítor Albiero and Vladan Petrovic and Weiwei Chu and Wenhan Xiong and Wenyin Fu and Whitney Meers and Xavier Martinet and Xiaodong Wang and Xiaofang Wang and Xiaoqing Ellen Tan and Xide Xia and Xinfeng Xie and Xuchao Jia and Xuewei Wang and Yaelle Goldschlag and Yashesh Gaur and Yasmine Babaei and Yi Wen and Yiwen Song and Yuchen Zhang and Yue Li and Yuning Mao and Zacharie Delpierre Coudert and Zheng Yan and Zhengxing Chen and Zoe Papakipos and Aaditya Singh and Aayushi Srivastava and Abha Jain and Adam Kelsey and Adam Shajnfeld and Adithya Gangidi and Adolfo Victoria and Ahuva Goldstand and Ajay Menon and Ajay Sharma and Alex Boesenberg and Alexei Baevski and Allie Feinstein and Amanda Kallet and Amit Sangani and Amos Teo and Anam Yunus and Andrei Lupu and Andres Alvarado and Andrew Caples and Andrew Gu and Andrew Ho and Andrew Poulton and Andrew Ryan and Ankit Ramchandani and Annie Dong and Annie Franco and Anuj Goyal and Aparajita Saraf and Arkabandhu Chowdhury and Ashley Gabriel and Ashwin Bharambe and Assaf Eisenman and Azadeh Yazdan and Beau James and Ben Maurer and Benjamin Leonhardi and Bernie Huang and Beth Loyd and Beto De Paola and Bhargavi Paranjape and Bing Liu and Bo Wu and Boyu Ni and Braden Hancock and Bram Wasti and Brandon Spence and Brani Stojkovic and Brian Gamido and Britt Montalvo and Carl Parker and Carly Burton and Catalina Mejia and Ce Liu and Changhan Wang and Changkyu Kim and Chao Zhou and Chester Hu and Ching-Hsiang Chu and Chris Cai and Chris Tindal and Christoph Feichtenhofer and Cynthia Gao and Damon Civin and Dana Beaty and Daniel Kreymer and Daniel Li and David Adkins and David Xu and Davide Testuggine and Delia David and Devi Parikh and Diana Liskovich and Didem Foss and Dingkang Wang and Duc Le and Dustin Holland and Edward Dowling and Eissa Jamil and Elaine Montgomery and Eleonora Presani and Emily Hahn and Emily Wood and Eric-Tuan Le and Erik Brinkman and Esteban Arcaute and Evan Dunbar and Evan Smothers and Fei Sun and Felix Kreuk and Feng Tian and Filippos Kokkinos and Firat Ozgenel and Francesco Caggioni and Frank Kanayet and Frank Seide and Gabriela Medina Florez and Gabriella Schwarz and Gada Badeer and Georgia Swee and Gil Halpern and Grant Herman and Grigory Sizov and Guangyi and Zhang and Guna Lakshminarayanan and Hakan Inan and Hamid Shojanazeri and Han Zou and Hannah Wang and Hanwen Zha and Haroun Habeeb and Harrison Rudolph and Helen Suk and Henry Aspegren and Hunter Goldman and Hongyuan Zhan and Ibrahim Damlaj and Igor Molybog and Igor Tufanov and Ilias Leontiadis and Irina-Elena Veliche and Itai Gat and Jake Weissman and James Geboski and James Kohli and Janice Lam and Japhet Asher and Jean-Baptiste Gaya and Jeff Marcus and Jeff Tang and Jennifer Chan and Jenny Zhen and Jeremy Reizenstein and Jeremy Teboul and Jessica Zhong and Jian Jin and Jingyi Yang and Joe Cummings and Jon Carvill and Jon Shepard and Jonathan McPhie and Jonathan Torres and Josh Ginsburg and Junjie Wang and Kai Wu and Kam Hou U and Karan Saxena and Kartikay Khandelwal and Katayoun Zand and Kathy Matosich and Kaushik Veeraraghavan and Kelly Michelena and Keqian Li and Kiran Jagadeesh and Kun Huang and Kunal Chawla and Kyle Huang and Lailin Chen and Lakshya Garg and Lavender A and Leandro Silva and Lee Bell and Lei Zhang and Liangpeng Guo and Licheng Yu and Liron Moshkovich and Luca Wehrstedt and Madian Khabsa and Manav Avalani and Manish Bhatt and Martynas Mankus and Matan Hasson and Matthew Lennie and Matthias Reso and Maxim Groshev and Maxim Naumov and Maya Lathi and Meghan Keneally and Miao Liu and Michael L. Seltzer and Michal Valko and Michelle Restrepo and Mihir Patel and Mik Vyatskov and Mikayel Samvelyan and Mike Clark and Mike Macey and Mike Wang and Miquel Jubert Hermoso and Mo Metanat and Mohammad Rastegari and Munish Bansal and Nandhini Santhanam and Natascha Parks and Natasha White and Navyata Bawa and Nayan Singhal and Nick Egebo and Nicolas Usunier and Nikhil Mehta and Nikolay Pavlovich Laptev and Ning Dong and Norman Cheng and Oleg Chernoguz and Olivia Hart and Omkar Salpekar and Ozlem Kalinli and Parkin Kent and Parth Parekh and Paul Saab and Pavan Balaji and Pedro Rittner and Philip Bontrager and Pierre Roux and Piotr Dollar and Polina Zvyagina and Prashant Ratanchandani and Pritish Yuvraj and Qian Liang and Rachad Alao and Rachel Rodriguez and Rafi Ayub and Raghotham Murthy and Raghu Nayani and Rahul Mitra and Rangaprabhu Parthasarathy and Raymond Li and Rebekkah Hogan and Robin Battey and Rocky Wang and Russ Howes and Ruty Rinott and Sachin Mehta and Sachin Siby and Sai Jayesh Bondu and Samyak Datta and Sara Chugh and Sara Hunt and Sargun Dhillon and Sasha Sidorov and Satadru Pan and Saurabh Mahajan and Saurabh Verma and Seiji Yamamoto and Sharadh Ramaswamy and Shaun Lindsay and Shaun Lindsay and Sheng Feng and Shenghao Lin and Shengxin Cindy Zha and Shishir Patil and Shiva Shankar and Shuqiang Zhang and Shuqiang Zhang and Sinong Wang and Sneha Agarwal and Soji Sajuyigbe and Soumith Chintala and Stephanie Max and Stephen Chen and Steve Kehoe and Steve Satterfield and Sudarshan Govindaprasad and Sumit Gupta and Summer Deng and Sungmin Cho and Sunny Virk and Suraj Subramanian and Sy Choudhury and Sydney Goldman and Tal Remez and Tamar Glaser and Tamara Best and Thilo Koehler and Thomas Robinson and Tianhe Li and Tianjun Zhang and Tim Matthews and Timothy Chou and Tzook Shaked and Varun Vontimitta and Victoria Ajayi and Victoria Montanez and Vijai Mohan and Vinay Satish Kumar and Vishal Mangla and Vlad Ionescu and Vlad Poenaru and Vlad Tiberiu Mihailescu and Vladimir Ivanov and Wei Li and Wenchen Wang and Wenwen Jiang and Wes Bouaziz and Will Constable and Xiaocheng Tang and Xiaojian Wu and Xiaolan Wang and Xilun Wu and Xinbo Gao and Yaniv Kleinman and Yanjun Chen and Ye Hu and Ye Jia and Ye Qi and Yenda Li and Yilin Zhang and Ying Zhang and Yossi Adi and Youngjin Nam and Yu and Wang and Yu Zhao and Yuchen Hao and Yundi Qian and Yunlu Li and Yuzi He and Zach Rait and Zachary DeVito and Zef Rosnbrick and Zhaoduo Wen and Zhenyu Yang and Zhiwei Zhao and Zhiyu Ma},
      year={2024},
      eprint={2407.21783},
      archivePrefix={arXiv},
      primaryClass={cs.AI},
      url={https://arxiv.org/abs/2407.21783}, 
}

@article{wirtz2025functional,
  title={Functional Prestige in Sociolinguistic Evaluative Judgements Among Adult Second Language Speakers in Austria: Evidence from Perception},
  author={Wirtz, Mason A and Ender, Andrea},
  journal={Languages},
  volume={10},
  number={4},
  pages={67},
  year={2025},
  publisher={MDPI}
}

@inbook{Rutten_Vosters_2021, place={Cambridge}, series={Cambridge Handbooks in Language and Linguistics}, title={Language Standardization ‘from Above’}, booktitle={The Cambridge Handbook of Language Standardization}, publisher={Cambridge University Press}, author={Rutten, Gijsbert and Vosters, Rik}, year={2021}, pages={65–92}, collection={Cambridge Handbooks in Language and Linguistics}}

@article{galla2016indigenous,
  title={Indigenous language revitalization, promotion, and education: Function of digital technology},
  author={Galla, Candace Kaleimamoowahinekapu},
  journal={Computer Assisted Language Learning},
  volume={29},
  number={7},
  pages={1137--1151},
  year={2016},
  publisher={Taylor \& Francis}
}

@inproceedings{mager-etal-2018-challenges,
    title = "Challenges of language technologies for the indigenous languages of the {A}mericas",
    author = "Mager, Manuel  and
      Gutierrez-Vasques, Ximena  and
      Sierra, Gerardo  and
      Meza-Ruiz, Ivan",
    editor = "Bender, Emily M.  and
      Derczynski, Leon  and
      Isabelle, Pierre",
    booktitle = "Proceedings of the 27th International Conference on Computational Linguistics",
    month = aug,
    year = "2018",
    address = "Santa Fe, New Mexico, USA",
    publisher = "Association for Computational Linguistics",
    url = "https://aclanthology.org/C18-1006/",
    pages = "55--69"
}

@inproceedings{liu-etal-2022-always,
    title = "Not always about you: Prioritizing community needs when developing endangered language technology",
    author = "Liu, Zoey  and
      Richardson, Crystal  and
      Hatcher, Richard  and
      Prud{'}hommeaux, Emily",
    editor = "Muresan, Smaranda  and
      Nakov, Preslav  and
      Villavicencio, Aline",
    booktitle = "Proceedings of the 60th Annual Meeting of the Association for Computational Linguistics (Volume 1: Long Papers)",
    month = may,
    year = "2022",
    address = "Dublin, Ireland",
    publisher = "Association for Computational Linguistics",
    url = "https://aclanthology.org/2022.acl-long.272/",
    doi = "10.18653/v1/2022.acl-long.272",
    pages = "3933--3944"
}

@inproceedings{mager-etal-2023-ethical,
    title = "Ethical Considerations for Machine Translation of Indigenous Languages: Giving a Voice to the Speakers",
    author = "Mager, Manuel  and
      Mager, Elisabeth  and
      Kann, Katharina  and
      Vu, Ngoc Thang",
    editor = "Rogers, Anna  and
      Boyd-Graber, Jordan  and
      Okazaki, Naoaki",
    booktitle = "Proceedings of the 61st Annual Meeting of the Association for Computational Linguistics (Volume 1: Long Papers)",
    month = jul,
    year = "2023",
    address = "Toronto, Canada",
    publisher = "Association for Computational Linguistics",
    url = "https://aclanthology.org/2023.acl-long.268/",
    doi = "10.18653/v1/2023.acl-long.268",
    pages = "4871--4897"
}

\appendix

\section{Code and Dataset} \label{ap:code}

We release our code at \url{https://github.com/MinhDucBui/Meenz-bleibt-Meenz}. Instructions for accessing the dataset are provided at the same link under the section ``Access to Dataset''. The dataset is distributed under the CC BY-NC-ND 4.0 license.

\section{Full Prompts} \label{ap:prompts}

We report the full prompts used in our study.

\subsection{Dataset Creation: Extracting Definition} \label{ap:prompt_dataset}

\begin{figure}[t]
    \centering
    \begin{promptbox}[Definition Extraction Prompt]
        \ttfamily\small\raggedright
        ``Du erhältst eine unstrukturierte oder fehlerhafte Definition des Wortes '\{Word\}' 
        aus einem alten Wörterbuch. Deine Aufgabe ist es, nur die eigentliche Bedeutung 
        des Wortes aus dem Text zu extrahieren, ohne Kommentare, Reformulierungen oder Erklärungen.\\[4pt]

        Regeln:\\
        1. Wenn mehrere Bedeutungen vorhanden sind, nummeriere sie fortlaufend (1., 2., 3., …) und trenne sie jeweils mit einem Zeilenumbruch.\\
        2. Wenn die Definition ausschließlich auf ein anderes Wort verweist, gib '[SIEHE] <Wort>' aus.\\
        3. Verändere den Originaltext nicht, sondern gib ausschließlich den relevanten Ausschnitt wieder.\\
        4. Wenn es keine Definition im Text gibt, gebe 'Keine Definition' wieder.\\
        <Few-Shot Examples>''
    \end{promptbox}
    \caption{\textbf{Prompt used for extracting dictionary definitions.} 
    \textit{English translation:} ``You receive an unstructured or faulty definition of the word '\{Word\}' from an old dictionary. Your task is to extract only the actual meaning of the word from the text, without comments, reformulations, or explanations. Rules: (1) If multiple meanings are present, number them consecutively (1., 2., 3., …) and separate them by line breaks. (2) If the definition refers exclusively to another word, output '[SEE] <word>'. (3) Do not alter the original text; output only the relevant excerpt. (4) If there is no definition in the text, output 'No definition'.''} 
    \label{fig:extraction_prompt}
\end{figure}

To extract and clean the dictionary definitions for the creation of our dataset (see Section~\ref{sec:dataset}), which constitutes the fifth step in our dataset construction pipeline, we report the full prompt in Figure~\ref{fig:extraction_prompt}. For brevity, we omit the system prompt (translated as: ``You are a precise and reliable assistant for extracting linguistic data. Provide exclusively the requested information without modifying, shortening, or adding content.''). Furthermore, the full prompt includes six in-context examples, which are documented in the code.

\begin{figure}
    \centering
    \begin{promptbox}[Definition Cleaning Prompt]
        \ttfamily\small\raggedright
        \{system\_token\}\\
        ``Du bist ein sorgfältiger linguistischer Assistent. 
        Deine Aufgabe ist es, Wörterbuchdefinitionen zu bereinigen, 
        ohne deren Bedeutung zu verändern. Entferne ausschließlich unnötige Sonderzeichen 
        wie Bindestriche, doppelte Leerzeichen oder ähnliche OCR-Artefakte. 
        Lasse Verweise in der Form [SIEHE] unverändert und Nummerierungen von Definitionen.''\\
        \{user\_token\}\\
        ``Hier ist die Definition für das Wort '\{Word\}':\\[4pt]
        \{cleaned\}\\[6pt]
        Bereinige die Definition gemäß den Anweisungen. 
        Wenn sie bereits sauber ist, gib sie unverändert zurück. 
        Gebe nur die bereinigte Definition wieder''
    \end{promptbox}
    \caption{\textbf{Prompt used for cleaning dictionary definitions}
    \textit{English translation:} ``You are a careful linguistic assistant. Your task is to clean dictionary definitions without changing their meaning. Remove only unnecessary special characters such as hyphens, double spaces, or similar OCR artifacts. Leave references in the form [SEE] unchanged and preserve numbering of definitions. Here is the definition for the word '\{Word\}': \{cleaned\}. Clean the definition according to the instructions. If it is already clean, return it unchanged. Output only the cleaned definition.''}
    \label{fig:cleaning_prompt}
\end{figure}

We clean each extracted definition of OCR-induced noise using Llama-3.3 70B; the corresponding prompt is shown in Figure~\ref{fig:cleaning_prompt}.

\subsection{Definition Generation} \label{ap:prompts_1}

Figure~\ref{fig:dictionary_prompt} presents the zero-shot prompt employed to generate dictionary definitions based on the Mainzer dialect in Section \ref{sec:understanding}. To generate the English word definitions for English words, we replace the ``Mainzer Dialekt'' part with ``Englisch'' and the ``Hochdeutsch'' (standard German) with ``Englisch''.

\begin{figure}[t]
    \centering
    \begin{promptbox}[Dictionary Definition Prompt]
        \ttfamily\small\raggedright
        \{system\_token\}\\
        ``Du bist ein präziser und zuverlässiger linguistischer Assistent. 
        Deine Aufgabe ist es, Wörterbuchdefinitionen zu erstellen. 
        Gib ausschließlich eine einzige, kurze und prägnante Bedeutung des angefragten Wortes an. 
        Nutze dabei die im Mainzer Dialekt gebräuchliche Bedeutung des Wortes, 
        formuliere die Definition jedoch auf Hochdeutsch.''\\
        \{user\_token\}\\
        ``Erstelle genau eine kurze Wörterbuchdefinition für das Wort '\{Word\}'. 
        Verwende dabei die im Mainzer Dialekt gebräuchliche Bedeutung, 
        ohne Zusatzinformationen, Beispiele oder alternative Bedeutungen anzugeben.''
    \end{promptbox}
    \caption{\textbf{Prompt used for generating dictionary definitions.} 
    \textit{English translation:} ``You are a precise and reliable linguistic assistant. Your task is to create dictionary definitions. Provide exclusively a single, short, and concise meaning of the requested word. Use the meaning commonly used in the Mainzer dialect, but formulate the definition in Standard German. Create exactly one short dictionary definition for the word '\{Word\}'. Use the meaning commonly used in the Mainzer dialect, without providing additional information, examples, or alternative meanings.''}
    \label{fig:dictionary_prompt}
\end{figure}

\begin{figure}[t]
    \centering
    \begin{promptbox}[Definition Similarity Evaluation Prompt]
        \ttfamily\small\raggedright
        \{system\_token\}\\
        ``Du bist ein präziser und neutraler Evaluator für Bedeutungsähnlichkeit von 
        Wörterbuchdefinitionen. 
        Deine Aufgabe ist es, zu beurteilen, ob zwei Definitionen inhaltlich dieselbe Bedeutung ausdrücken, 
        auch wenn sie unterschiedlich formuliert sind. 
        Berücksichtige Synonyme, Umformulierungen oder stilistische Unterschiede. 
        Antworte ausschließlich mit 'GLEICH', wenn die beiden Definitionen denselben Inhalt wiedergeben, 
        oder mit 'UNTERSCHIEDLICH', wenn sich ihre Bedeutung wesentlich unterscheidet. 
        Wörterbucheinträge können mehrere Definitionen enthalten. Wenn mindestens eine dieser Definitionen mit der vom LLM erzeugten übereinstimmt, gilt das Ergebnis als 'GLEICH'. 
        Gib keine weiteren Erklärungen oder Begründungen.''\\
        \{user\_token\}\\
        ``Vergleiche die folgenden zwei Definitionen. 
        Beurteile, ob sie inhaltlich gleich sind.\\[4pt]
        Definition A (LLM): '\{response\_generator\}'\\[4pt]
        Definition B (Wörterbuch): '\{definition\_final\}'''
    \end{promptbox}
    \caption{\textbf{Prompt used for evaluating semantic equivalence.} 
    \textit{English translation:} ``You are a precise and neutral evaluator of semantic similarity between dictionary definitions. Your task is to assess whether two definitions express the same meaning, even if they are phrased differently. Consider synonyms, paraphrases, or stylistic differences. Respond exclusively with 'SAME' if the two definitions convey the same content, or with 'DIFFERENT' if their meaning differs substantially. Dictionary entries may contain multiple definitions. If at least one of these definitions matches the one generated by the LLM, the result counts as 'SAME'. Do not provide any additional explanations or justifications. Compare the following two definitions. Assess whether they are semantically equivalent. Definition A (LLM): '\{response\_generator\}'. Definition B (Dictionary): '\{definition\_final\}'.''}
    \label{fig:similarity_prompt}
\end{figure}

The complete prompt used for the LLM-as-a-Judge (LLMaaJ) approach is shown in Figure~\ref{fig:similarity_prompt}.

\subsection{Dialect Word Generation} \label{ap:prompts_2}

We report the word-generation prompt from Section~\ref{sec:generation} in Figure~\ref{fig:dialect_word_prompt}. For the English setup, we simply replace ``Mainzer Dialekt'' with ``Englisch'' and ``Hochdeutsch'' (standard German) with ``Englisch''.

\begin{figure}[t]
    \centering
    \begin{promptbox}[Dialect Word Generation Prompt]
        \ttfamily\small\raggedright
        \{system\_token\}\\
        ``Du bist ein präziser und zuverlässiger linguistischer Assistent. 
        Deine Aufgabe ist es, zu einem gegebenen Bedeutungsinhalt das passende Wort 
        im Mainzer Dialekt zu finden. 
        Gib ausschließlich ein einziges Wort aus, das diese Bedeutung im Mainzer Dialekt ausdrückt. 
        Gib keine Erklärungen, Übersetzungen oder Zusatzinformationen.''\\
        \{user\_token\}\\
        ``Finde das Mainzer Dialektwort, das die folgende Bedeutung ausdrückt:\\[4pt]
        '\{definition\_final\_generation\}'\\[4pt]
        Antworte nur mit dem Dialektwort.''
    \end{promptbox}
    \caption{\textbf{Prompt used for generating a Mainzer dialect word}. 
    \textit{English translation:} ``You are a precise and reliable linguistic assistant. Your task is to find the appropriate word in the Mainzer dialect for a given meaning. Provide exclusively a single word that expresses this meaning in the Mainzer dialect. Do not provide explanations, translations, or additional information. Find the Mainzer dialect word that expresses the following meaning: '\{definition\_final\_generation\}'. Respond only with the dialect word.''}
    \label{fig:dialect_word_prompt}
\end{figure}

\subsection{Few-Shot Learning}
For the few-shot setting, we augment the prompt shown in Figure~\ref{fig:dictionary_prompt} by appending examples in the following format: ``Wort: <WORD> Definition: <DEFINITION>''.

\subsection{Automatic Rule Generation} \label{sec:automatic_rule}

\begin{figure}[t]
    \centering
    \begin{promptbox}[Two-Turn Prompt for Rule Extraction]
        \ttfamily\small\raggedright
        
        ----------- Turn 1 -----------\\[6pt]
        
        Given the following CSV table, you need to perform the following tasks:\\
        Create an additional column with a German word. The word should fit the definition and be as similar as possible to the ``Word''.\\[6pt]
        
        <TRAINING SET>\\[10pt]
        
        ------- Turn 2 -------\\[6pt]
        
        Now, try to find common mappings between the dialect words and the German words. The mapping might not apply to all cases, but attempt to identify recurring similarities and patterns.\\
        
        Write explicit rules and pattern descriptions that could be used by a human. The rules should be concrete but comprehensive.
        
    \end{promptbox}
    \caption{\textbf{Two-turn prompting strategy for automatic rule extraction.} In the first turn, the model generates Standard German candidates. In the second turn, it induces mapping rules between dialect and Standard German forms.}
    \label{fig:two_turn_prompt}
\end{figure}

We first present the prompts used for automatic rule generation (see Figure~\ref{fig:two_turn_prompt}). We adopt a two-turn approach: in the first turn, the model generates Standard German candidates; in the second turn, it induces mapping rules between the dialect and Standard German forms.

The complete set of extracted rules, described in Section~\ref{sec:automatic_rule_generation}, is provided in Figure~\ref{fig:mapping_rules}.

\begin{figure}[t]
    \centering
    \begin{promptbox}[Dialect-to-Standard Mapping Rules]
        \ttfamily\small\raggedright
        \#\# Practical Application Rules\\[6pt]

        \#\#\# Step-by-Step Mapping Process\\[4pt]

        1. Identify Core Root\\
        - Extract the main semantic element from the dialect word\\
        - Example: `Klebberschulde` → focus on `Klebb` and `Schuld`\\[4pt]

        2. Apply Sound Transformations\\
        - Apply systematic sound changes based on patterns above\\
        - Example: `Klebb` → `Kleb` (consonant simplification)\\[4pt]

        3. Handle Suffixes Systematically\\
        - Apply standard German suffix equivalents\\
        - Example: `-schulde` → `-schulden` (noun pluralization)\\[4pt]

        4. Consider Semantic Context\\
        - Use definition to guide final word choice\\
        - Ensure the mapped word fits the semantic field\\[8pt]

        \#\#\# Quick Reference Guide\\[4pt]

        | Dialect Pattern | Standard German Equivalent | Example |\\
        |----------------|---------------------------|---------|\\
        | `-che` | `-chen` | `Bobbelche` → `Bobby` |\\
        | `-ele` | `-eln` | `knerchele` → `knirschen` |\\
        | `aa` | `ei` | `Gaawer` → `Geifer` |\\
        | Final `-e` | Often added to nouns | `Grobbe` → `Grob` |\\
        | Compound words | Preserve first element | `Klebberschulde` → `Kleberschulden` |\\
        | Professional `-er` | Usually preserved | `Stinkerd` → `Stinker` |
    \end{promptbox}
    \caption{\textbf{Automatic extracted rules for mapping Mainzer dialect words to Standard German forms.} This is generated by DeepSeek-R1.}
    \label{fig:mapping_rules}
\end{figure}

\end{document}